\documentclass[letterpaper]{article} % DO NOT CHANGE THIS
\usepackage[]{aaai2026}  % DO NOT CHANGE THIS
\usepackage{aaai2026}  % DO NOT CHANGE THIS
\usepackage{times}  % DO NOT CHANGE THIS
\usepackage{helvet}  % DO NOT CHANGE THIS
\usepackage{courier}  % DO NOT CHANGE THIS
\usepackage[hyphens]{url}  % DO NOT CHANGE THIS
\usepackage{graphicx} % DO NOT CHANGE THIS
\usepackage{adjustbox}
\usepackage{natbib}  % DO NOT CHANGE THIS AND DO NOT ADD ANY OPTIONS TO IT
\usepackage{caption} % DO NOT CHANGE THIS AND DO NOT ADD ANY OPTIONS TO IT
\usepackage[ruled,vlined,algo2e]{algorithm2e} % Added for \KwIn and algorithm environment support
\usepackage{algorithm}

\usepackage{newfloat}
\usepackage{listings}

\DeclareCaptionStyle{ruled}{labelfont=normalfont,labelsep=colon,strut=off} % DO NOT CHANGE THIS
\floatstyle{ruled}
\newfloat{listing}{tb}{lst}{}
\floatname{listing}{Listing}
\title{10 Open Challenges Steering the Future of Vision-Language-Action Models}
\author{
    Soujanya Poria\textsuperscript{\rm 1},
    Navonil Majumder\textsuperscript{\rm 2},
    Chia-Yu Hung\textsuperscript{\rm 1},
    Amir Ali Bagherzadeh\textsuperscript{\rm 3},
    Chuan Li\textsuperscript{\rm 3},\\
    Kenneth Kwok\textsuperscript{\rm 5},
    Ziwei Wang\textsuperscript{\rm 1},
    Cheston Tan\textsuperscript{\rm 5},
    Jiajun Wu\textsuperscript{\rm 6},
    David Hsu\textsuperscript{\rm 4}
}
\affiliations{
    \textsuperscript{\rm 1}Nanyang Technological University, 
    \textsuperscript{\rm 2}SUTD,
    \textsuperscript{\rm 3}Lambda Labs,
    \textsuperscript{\rm 4}National University of Singapore\\
    \textsuperscript{\rm 5}A*STAR,
    \textsuperscript{\rm 6}Stanford University\\
}

\usepackage{graphicx}
\usepackage{caption}
\usepackage{algorithm}
\usepackage{algorithmic}
\usepackage{multirow}
\usepackage{amsfonts,amsmath,amssymb}
\usepackage{subcaption}
\usepackage[capitalize]{cleveref}
\usepackage{enumitem}
\usepackage{pifont}
\usepackage{newfloat}
\usepackage{listings}
\usepackage{soul}
\usepackage[table,xcdraw]{xcolor}
\usepackage{tikz}
\usepackage{arydshln}
\usepackage{tabularx}
\usepackage{booktabs}
\usepackage{tcolorbox}
\usepackage{mdframed}
\usepackage{fancyvrb}
\usepackage{adjustbox}
\usepackage{mdframed}
\definecolor{lightyellow}{HTML}{ffe599}
\definecolor{green}{HTML}{34a853}
\definecolor{lightcornflowerblue}{HTML}{c9daf8}
\definecolor{darkyellow}{rgb}{0.85, 0.65, 0.13}

\definecolor{nmcolor}{RGB}{255, 25, 26}

\definecolor{jamcolor}{HTML}{005F73} % Teal-ish color
\definecolor{linkbg}{HTML}{F1FAEE}   % Soft background
\definecolor{lightergray}{RGB}{217, 219, 221}
\definecolor{psy}{RGB}{250, 230, 254}

\begin{document}

\maketitle

\begin{figure}
  \centering
  \includegraphics[width=0.9\linewidth]{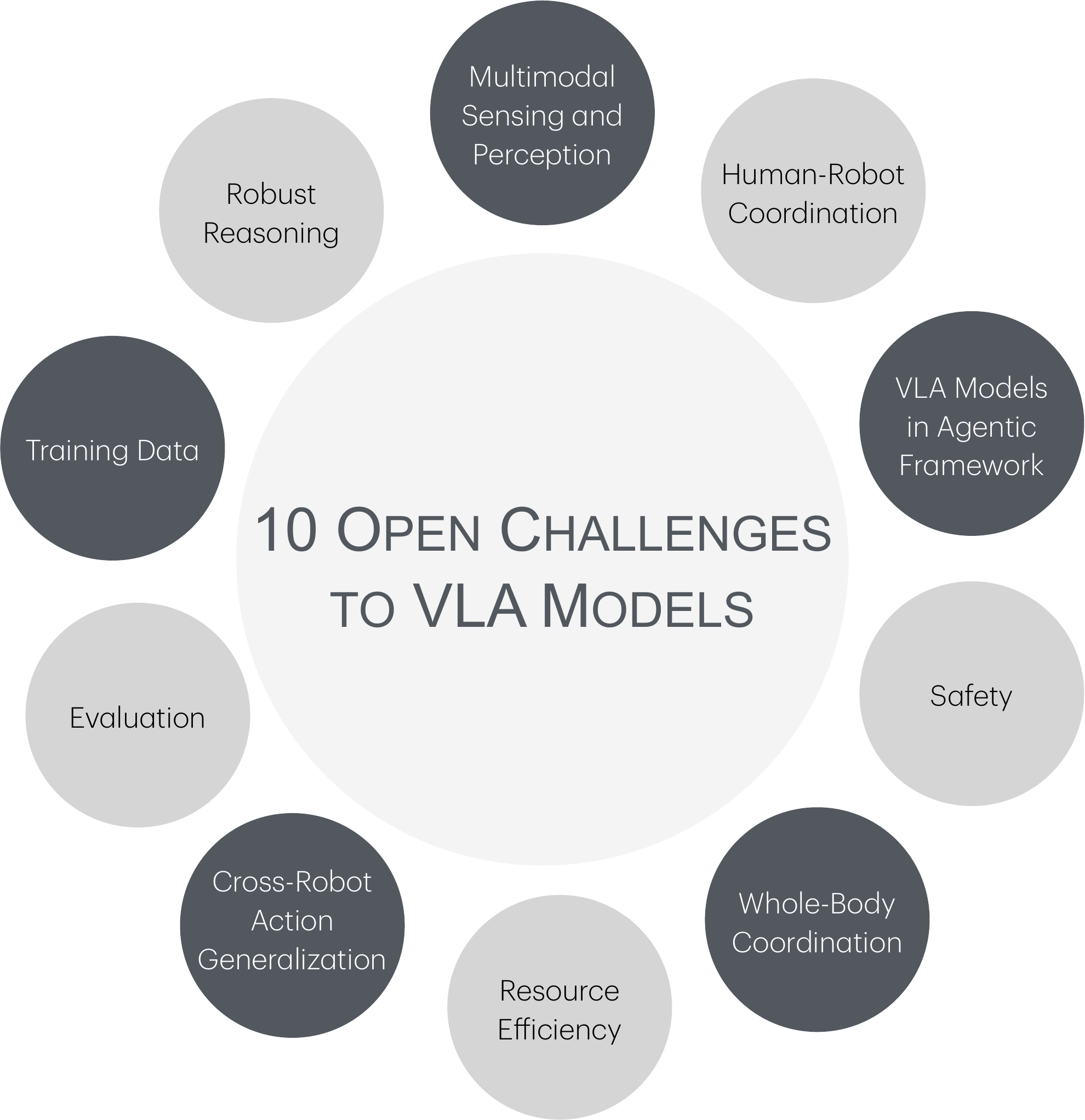}
\end{figure}

\begin{abstract}
Due to their ability of follow natural language instructions, vision-language-action (VLA) models are increasingly prevalent in the embodied AI arena, following the widespread success of their precursors---LLMs and VLMs. In this paper, we discuss 10 principal milestones in the ongoing development of VLA models---multimodality, reasoning, data, evaluation, cross-robot action generalization, efficiency, whole-body coordination, safety, agents, and coordination with humans. Furthermore, we discuss the emerging trends of using spatial understanding, modeling world dynamics, post training, and data synthesis---all aiming to reach these milestones. Through these discussions, we hope to bring attention to the research avenues that may accelerate the development of VLA models into wider acceptability.
\end{abstract}

\section{Introduction to VLA Models}
Enabling robots to perform a wide range of manipulation tasks in unstructured real-world environments is a central goal of modern robotics.
Robotic policy models, which aims to generate sequences of low-level actions (e.g., end-effector poses or joint torques) for complex tasks, have been at the forefront of this effort. Traditional reinforcement learning-based approaches have achieved impressive results on narrowly-defined tasks within fixed environments \citep{ma2024survey}, but they often struggle to generalize beyond their training scenarios \citep{brohan2023rt1roboticstransformerrealworld,chi2023diffusion}. The immense diversity and complexity of real-world scenarios pose a significant challenge for robotic policies, as they must possess strong generalization capabilities to perform effectively across multiple tasks and environments in the real world.

Recent progress in the foundation models for vision and language has demonstrated remarkable capabilities in scene understanding and task planning \citep{radford2021learning,zhai2023sigmoid,touvron2023llama}. Visual-Language Models (VLMs) have emerged as powerful tools for interpreting and reasoning about the visual world, demonstrating proficiency in tasks from object detection to video understanding. Their capacity to use chain-of-thought to decompose complex visual or language tasks into a sequence of logical steps makes them a promising platform for high-level robotic planning. Yet, a fundamental gap remains: VLMs are not designed and trained to produce the robot-executable policies needed to navigate the specific physical constraints and control dynamics of a given robot.

To bridge this gap, \textbf{Visual-Language-Action (VLA) models} have emerged that combine visual observations and language instructions to produce generalized robotic actions across diverse environments. The tasks are learned through Imitation Learning (IL), a paradigm where a robot learns to execute tasks by observing and replicating expert demonstrations. This approach allows the policy to generalize across a wide range of environments and tasks. These models can be broadly categorized into two groups, based on their action modeling strategy:
\begin{itemize}[itemsep=0pt, leftmargin=*, wide, labelwidth=0pt, labelindent=0pt, parsep=0pt, topsep=0pt]
    \item \textbf{Continuous-action models} \cite{octo_2023}, which typically leverage diffusion processes to generate smooth trajectories in continuous action spaces. 
    \item \textbf{Discrete-action models} \cite{brohan2023rt, brohan2023rt1roboticstransformerrealworld, kim2024openvla, sun2024emmaxembodiedmultimodalaction}, where robot actions are represented as sequences of discrete tokens. Due to compression through quantization, this approach risks lossy tokenization and detokenization.
    % , typically generated by transformers through autoregression.
\end{itemize}

In the standard VLA setting for imitation learning, the robot's state at time $t$ is described by a multimodal observation $(O_t, L_t, S_t)$, including current observation $O_t$, textual instructions $L_t$, and prior context $S_t$. 
The objective is to maximize the log probability of a sequence of actions $A_t$ generated by the equation
$
\max_{\theta} \mathbb{E}_{(A_{t}, O_t, L_t,S_t) \sim \mathcal{D}} \log \pi_{\theta}(A_{t} | O_t, L_t,S_t)$, where D is the demonstration dataset.$
\pi_{\theta}(A_t \mid O_t, L_t, S_t)
$. For cases with discrete action representations, this can be framed as a standard next-token prediction problem over a sequence of action tokens, a straightforward approach analogous to autoregressive language modeling. Another family of approaches utilizes diffusion \cite{ho2020denoisingdiffusionprobabilisticmodels} or flow matching \cite{lipman2023flowmatchinggenerativemodeling} to model the continuous action distribution \cite{octo_2023,chi2023diffusionpolicy}. This dichotomy has led to two primary schools of thought for modeling the action space: discrete action representation and continuous action representation, each with distinct trade-offs. 
Discrete action models can be easily implemented with transformer-based language models, allowing them to leverage next-token prediction. However, this approach introduces quantization errors and its auto-regression incurs slow inference speeds, making it unsuitable for high-frequency control \cite{kim2025finetuningvisionlanguageactionmodelsoptimizing}. In contrast, diffusion-based continuous action models jointly predict an action chunk, better preserving the fidelity of robotic movements and better suiting high-frequency control, but requires a significantly higher compute budget for convergence. 

\subsection{Discrete Action Models}
\label{sec:act-tok}

Robot actions are inherently continuous, spanning multiple degrees of freedom such as 3D translation $(x, y, z)$ and rotation (roll, pitch, yaw). To integrate these actions with transformer-based language models, it is common to discretize them into finite bins \cite{brohan2023rt1roboticstransformerrealworld, brohan2023rt}. A typical approach maps each action dimension to one of 256 discrete bins using a quantile-based strategy, which balances granularity with robustness to outliers. 

OpenVLA \cite{kim2024openvla} incorporates these action tokens directly into the language model vocabulary by replacing the 256 least-used tokens of the LLaMA tokenizer. This enables the model to perform next-token prediction over action sequences in the same way it predicts text. To further improve efficiency, the FAST+ tokenization method ~\cite{pertsch2025fastefficientactiontokenization} applies a discrete cosine transform (DCT) across action dimensions at each timestep, decorrelating joint action components. The resulting sequences can then be compressed using byte-pair encoding (BPE), reducing vocabulary size and sequence length while preserving essential structure. This allows robot action to be represented with a few tokens, which accelerates training and inference.

\subsection{Continuous Action Models}
In contrast to discrete action tokenization, continuous-action models directly parameterize and predict a more expressive continuous action distribution. This approach aims to preserve the inherent fidelity of robotic movements, avoiding the quantization errors that can arise from binning. Discrete-Action Models are also unsuitable for high-frequency controls due to slow inference speed (3-5Hz) ~\cite{kim2025finetuningvisionlanguageactionmodelsoptimizing} and unreliable task execution on bimanual manipulators. Prominent approaches ~\cite{pi02024,intelligence2025pi05visionlanguageactionmodelopenworld} add an additional small action expert that models the continuous actions conditioned on the internal hidden states of the larger VLM, effectively decoupling the action generation task from the VLM's primary function of multimodal representation. However, a key limitation is that training diffusion policies requires a significantly higher computational budget for convergence~\cite{pertsch2025fastefficientactiontokenization}. This disparity has motivated the development of hybrid approaches that combine the strengths of both paradigms. Specifically, VLA models are first pretrained autoregressively on discrete action tokens, after which a specialized action expert is integrated and further trained to produce continuous action outputs. This methodology has demonstrated superior performance and significantly faster convergence ~\cite{intelligence2025pi05visionlanguageactionmodelopenworld}. Furthermore, to improve the generalization capabilities of the model, knowledge insulation ~\cite{driess2025knowledgeinsulatingvisionlanguageactionmodels} was introduced to preserve the core semantic knowledge of the VLM while training the action expert.

\section{The 10 Open Challenges}
Despite this great progress in this research field, there remain certain challenges that we discuss below.

\subsection{Multimodal Sensing and Perception}
\label{sec:chal-perc}

\paragraph{Depth Perception.} Many VLA models, except MolmoAct~\cite{lee2025molmoactactionreasoningmodels} and SpatialVLA~\cite{spatialvla2025}, ignore explicit depth information to robustly operate in a 3D environment. While depth perception could be approximated through comparing visual frames across time, such an approximation loses sensitivity with the object size, distance from the camera, and fineness of movements, which may precipitate into erroneous reasoning by the VLA.

Even though MolmoAct and SpatialVLA uses depth information, it is limited to the training phase where the VLA explicitly learns to impute depth from the RGB frames. This naturally eliminates the need for a specialized depth-gauging camera, but the limitation still remains due to the above reasons.

\paragraph{Environmental Noise and Artifacts.} Current evaluations of VLA models are evaluated under very controlled environments, be it in real world or simulation. To this end, SimplerEval~\cite{li24simpler} applies distribution shift across background, lighting, distractors, texture, and camera pose. However, a broad range of noise and artifacts, such as, reflections, lens flares, environment-induced noisy camera feed---due to water, dust, or debris---,  are still unaccounted for in the current evaluation frameworks.

\paragraph{Beyond Vision Modality.} At this nascent stage, VLA models, as named, are limited to visual and language modalities. However, there is strong incentive to expand the domain of these models into audio, speech, touch among other sensory information. For instance, embodied AI systems equipped with such multimodal action models could be deployed into hazardous disaster zones for rescue operations---audio signal could be instrumental here to listen to victims' call for help, respond to explosive sounds or falling debris, etc. On the other hand, touch modality would allow VLA to perform delicate tasks that require careful application of force through the joints and gripper---handling of glass and ceramic items, cooking, assemble and disassembly of electronic and electrical devices, etc. Recently, \citet{bi2025vlatouchenhancingvisionlanguageactionmodels} proposed a touch capable VLA that has shown to enhance overall performance on various pick, place, wipe tasks over vision-only VLA models.

\subsection{Robust Reasoning}
\label{sec:chal-reas}

Owing to being pre-trained on a large amount of text data and subsequent large-scale post-training, LLMs~\cite{bai2023qwen,openaiGPT4TechnicalReport2023,touvronLLaMAOpenEfficient2023} and, by extension, VLMs~\cite{Qwen2.5-VL,chen2023pali3,openaiGPT4TechnicalReport2023} are reasonably adept at high-level reasoning and planning to solve complex problems. However, such reasoning performance do not translate that well into the derived VLA models on seemingly much simpler robot tasks. The recent VLA models, such as, Emma-X~\cite{sun2024emmaxembodiedmultimodalaction}, CoT-VLA~\cite{cotvla2025}, and MolmoAct~\cite{lee2025molmoactactionreasoningmodels}, that are trained on both high-level linguistic and low-level action-level reasoning traces still produce imperfect results in both real-life and simulations, such as, LIBERO~\cite{liu2024libero} and SimplerEnv~\cite{li24simpler}. These evaluations usually involve simple tasks like picking and placing items, operating a drawer, pointing to an item, etc. To be deployable in more complex and sensitive environments, the error rate on such simpler tasks must approach near perfection. Robust reasoning becomes even more important for long-horizon tasks where the performance generally drops with increasing horizon. Furthermore, robust chain-of-thought reasoning has been shown to improve LLM performance on out-of-distribution tasks, which is also shown to be the trend for VLA models like Emma-X, CoT-VLA, and MolmoAct.

Another key challenge lies in the effective use of tools. In everyday life, even simple goals—such as making coffee—require identifying and using the right tools (e.g., a cup, a spoon, and so on). Equipping VLAs with the ability to understand, select, and utilize tools in a similar manner remains an important open problem.

\subsection{Quality Training Data}
\label{sec:chal-data}

Open-X-Embodiment~\cite{open_x_embodiment_rt_x_2023} is a unification of around 70 and growing smaller datasets that cover more than 1M episodes of diverse robots performing various tasks involving many items and scenes. Despite being trained on such a vast dataset, VLA models are often brittle to out-of-distribution environments and robot setups, requiring the collection of additional episodes for fine-tuning. 

There have also been attempts at using simulations to collect trajectories at a cheaper rate, like the Sim2Real~\cite{Kadian_2020} challenge, where models are trained on simulation data and evaluated on real robots. However, constructing real-enough simulations has been a challenge for both data collection and evaluation.

Finally, a major challenge arises from the variance and noise in the collected data~\cite{zheng2025universal}. Such variability can stem from differences in embodiments, camera positions and angles, as well as inconsistencies in the cognitive behaviors of human data collectors. 

\subsection{Evaluation of VLA Models}
\label{sec:chal-eval}

Evaluation of VLA models is a great challenge due to general limited availability of robots, environments, and objects. Thus, most VLA models are evaluated on a few real benchmarks that involve WidowX~\cite{sun2024emmaxembodiedmultimodalaction,kim2024openvla} or Franka~\cite{kim2024openvla} robots in a predefined environment, such as a kitchen with related objects therein. However, such evaluations obviously do not capture the overall capability of the VLA models spanning a broad range of robots and settings. Simulation-based evaluations~\cite{li24simpler,liu2023libero} somewhat mitigate this issue by evaluating the models on a finite but much larger set of environments with variable texture, lighting, camera pose, backdrop, etc.

However, the environments simulated in such tools often fail to capture enough details of their real-life counterparts, leading to poor correlation between in-simulation and real-life performance. The gap could be in the environmental details, such as, lighting, reflection, texture and consistency of the object surfaces, as well as the PD parameters (stiffness and damping parameters) of the various robot joints and gripper.  SimplerEnv~\cite{li24simpler} made great strides in minimizing this distribution shift in the robot motion through simulated annealing and in environments with image in-painting. However, there still remains much room for improvement, not only in terms of environmental and embodiment diversity and fidelity, but also incorporation of modalities, such as, audio and touch that have lately been gaining much research interest.

\subsection{Cross-Robot Action Generalization}
\label{sec:chal-gen}

Cross-robot action generalization remains a fundamental challenge for VLA models, primarily due to action heterogeneity. \citet{zheng2025universal} shows that training on action data from a fixed set of embodiments often fails to generalize to others with distinct action spaces---for example, robots with higher degree of freedom or structural differences across robotic arms, quadrupeds, and autonomous vehicles. The diversity of control interfaces further exacerbates this issue. Addressing action heterogeneity is therefore a critical first step toward building VLA models that can adapt seamlessly and achieve zero-shot generalization, analogous to the generality demonstrated by VLMs and LLMs in multimodal tasks.

\subsection{Resource Efficiency}
\label{sec:chal-res}

Embodiments/robots are generally much more compute-limited than the training infrastructure due to the constraints on space and energy requirements. To circumvent this limitation, the embodiments often act as thin clients that collect input and observations from the environment and relay them to a server with ample compute to infer the next actions from a VLA model. However, this setting is limited by the network communication latency and disruption---for instance, disasters zones are often cutoff from the internet and telecommunication services. Thus, embodiments are equipped with smaller and more resource efficient VLA models capable of running on the hardware on-board. Unfortunately, these smaller models are generally less performant as compared to their larger counterparts~\cite{octo_2023,kim2024openvla,rt22023arxiv,lee2025molmoactactionreasoningmodels}. Thus, striking the right balance between VLA model capacity and resource efficiency remains a key hurdle in the adoption of VLA models.

\subsection{Whole-Body Coordination}
\label{sec:chal-wb-coord}
Real-world VLA tasks often require an agent’s entire body to act in concert.
A mobile manipulator may reposition its base while moving its arm to reach, grasp, or carry objects in a safe manner. Enabling such whole-body coordination by coupling locomotion and manipulation under uncertainty is a core challenge for embodied VLA agents. Broadly, two approaches exist: (1) model-based control and (2) learning-based control. Model-based controllers, such as Model Predictive Control (MPC), optimize short-horizon trajectories for base and arm subject to dynamic balance, contact feasibility, and joint/torque limits~\cite{lecleach2024fastcimpc, kim2025cimpc}. By explicitly modeling robot dynamics, MPC yields precise, interpretable coordination and safety guarantees in structured settings. However, it relies on accurate models, faces real-time bottlenecks as dimensionality and non-smooth contact events grow, and often requires embodiment-specific tuning, complicating generalization and integration with high-level vision–language plans \cite{wang2025wbmpc, heins2023uprightmpc}. Data-driven policies learn unified locomotion–manipulation behaviors from multimodal observations (e.g., vision, proprioception, force/tactile) and demonstration~\cite{fu2024mobilealoha, ha2024umionlegs, chen2025acdit}. Reinforcement and imitation learning can decide when to move the base versus the arm, and recent systems expand the action space to include base motion alongside end-effector control. While adaptable to unstructured settings, these methods struggle with exploration and credit assignment, require careful reward shaping, and may generalize poorly across embodiments~\cite{liu2024visual, ha2024umionlegs}. Future progress will likely hinge on hybrid control frameworks that combine analytical structure with learning-based flexibility—for example, using constraint-aware planners to ensure feasibility and safety while learned policies provide goal-directed coordination, or learning to tune costs and propose feasible whole-body poses~\cite{zheng2025localreactivewbc, xie2025kungfubot, ji2024exbody}. A central need is objective and reward design that explicitly couples locomotion with manipulation so base motion contributes to stable, precise end-effector control~\cite{sundaresan2025homer, dugar2024mmwbc, zhao2025reactivewbcloco, du2024efficientwbmpc, xie2025kungfubot}. Ultimately, the dominant challenge is the high-dimensional search space of coupled actions, motivating compact action representations and hierarchical planning.

\subsection{Safety Assurances}
\label{sec:chal-safety}

The recent proliferation and wide accessibility of powerful LLMs like GPT, Claude, and Gemini have raised alarm bells across many institutions, from education and law to healthcare and security. One of the major concern is the LLMs producing harmful responses, truthful or otherwise, having negative influence on the users or the society as a whole. Such concern translates to the embodied AI systems as well, except the harm could be inflicted very directly through the actions, without any direct human supervision. For instance, in a rescue operation in a disaster zone, a robot equipped with embodied AI may harm the victims while saving them due to imperfect actions. To avoid such scenarios that may harm people and the overall trust in embodied AI systems, the community must devise appropriate guardrails and fail-safes. To this end, \citet{zhang25safevla} recently proposed an reinforcement learning-based approach for safety alignment of VLA models by constraining the actions while maintaining general performance.

\subsection{VLA Models in Agentic Frameworks}
\label{sec:chal-agent}
Human achievements are often made possible through collaboration in groups. Inspired by this, Agentic AI is a powerful framework increasingly relevant due to their general ability to solve problems through inter-agent communication, without being constrained to the local tools available to an individual agent. Such multi-agent framework is yet to be thoroughly explored through the lens of VLA models. Adoption of such framework could address the agent-level resource constraints discussed in \cref{sec:chal-res}. For instance, certain compute load can be delegated to a nearby idle robot. On the other hand, another robot equipped with certain special sensor or at a different vantage point could collect observations for the benefit of another robot for improved decision making.
VLA models capable of such inter-agent communication could be instrumental to the applicability and growth of embodied AI systems. At the same time, an important challenge lies in determining the appropriate level of agency and autonomy to assign to each agent. More broadly, realizing such multi-agent VLA systems requires the development of trustworthy, safe, and verifiable workflow generation to reliably execute complex tasks. 

Recently \citep{yang2025agenticrobotbraininspiredframework} proposed a heterogeneous agentic framework of VLM/LLM/VLA models that consists of an LLM as a high-level planner and another VLM as a verifier to assist the VLA model carry out the given tasks. This early work shows a promising path toward truly-embodied VLA agents.

\subsection{Human-Robot Coordination}
\label{sec:chal-comm}

The explicit communication between users and current VLA models is unidirectional, i.e., human to robot through natural language instructions. Going forward, natural language and visual output---containing reasoning traces and questions for the user seeking missing information---from the VLA models could be fruitful for interpretation of the actions and user interaction. Recently, CoT-VLA~\cite{cotvla2025} has been trained to produce a visual output of the intended state prior to action decoding; this has been shown to be improving the overall model performance across the standard benchmarks. Emma-X~\cite{sun2024emmaxembodiedmultimodalaction} trained to generate high-level rationale in natural language, has shown to outperform direct action generation.

\section{Emerging Trends to Face the Challenges}
We cover the emerging trends, pivotal to the progress of VLA models.

\begin{figure*}
    \centering
    \includegraphics[width=\linewidth]{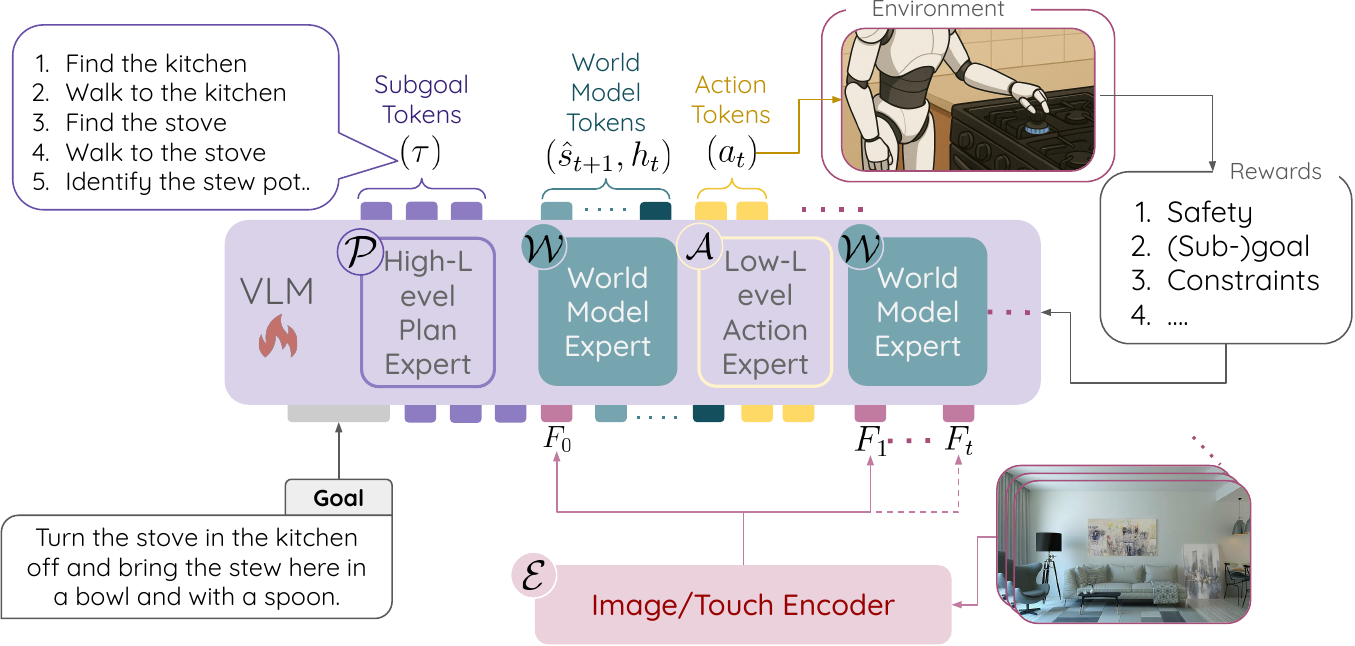}
    \caption{A high-level emerging VLA framework.}
    \label{fig:framework}
\end{figure*}

\subsection{Hierarchical Planning} 
We cast the class of Embodied‐AI tasks addressed by VLA models as hierarchical planning problems that could be addressed by the following high-level framework (see \cref{fig:framework}):
%\begin{mdframed}[backgroundcolor=green!5]
\begin{algorithm}[ht]
\footnotesize
\caption{Multi-Agent VLA Planning with Orchestrator, Workflow Generation, and Safety Guardrails}
\label{alg:multiagent-vla}
\KwIn{
  Initial observation/state $s_0$ (e.g., image, joint readings, etc.); \\
  Goal specification $G$ in natural language; \\
  Library of expert skill policies $\{\pi_1,\dots,\pi_K\}$; \\
  High-level planner (also serving as orchestrator and workflow generator).
}
\KwOut{Successful execution achieving goal $G$ if $s^{(N)} \models G$.}

\BlankLine
\textbf{Step 1: High-level orchestration and workflow generation} \\
The high-level planner/orchestrator decomposes $G$ into a workflow of subtasks:
\[
  P = [\,\tau^{(1)}, \tau^{(2)}, \dots, \tau^{(N)}\,],
\]
where each $\tau^{(i)}$ is assigned to an expert VLA specialized in that skill.

\BlankLine
\textbf{Step 2: Subgoal execution with expert VLAs} \\
\For{$i=1$ \KwTo $N$}{
  Each subtask $\tau^{(i)}$ is executed as a sequence of low-level actions:
  \[
    A^{(i)} = [a^{(i)}_{1}, a^{(i)}_{2}, \dots, a^{(i)}_{T_i}],
  \]
  where for each $j=1,\dots,T_i$:
  \[
    a^{(i)}_{j} \sim \pi_{k_i}\bigl(s_{t_j}^{(i-1)}, \tau^{(i)}\bigr),
    \quad
    s_{t_j+1}^{(i-1)} = \mathcal{W}(s_{t_j}^{(i-1)}, a^{(i)}_{j}).
  \]
  Execution yields state $s^{(i)}$ after satisfying $\tau^{(i)}$.
  
  \BlankLine
  \textbf{Step 2.1: Safety guardrails and evaluators} \\
  After completion of $A^{(i)}$, invoke safety checks through world model simulation $\mathcal{W}$ and evaluators. If feedback indicates plan inconsistency or safety risks, trigger \textbf{re-planning} by returning control to Step 1.
}

\BlankLine
\textbf{Step 3: Coordination} \\
The resulting $s^{(i)}$ becomes the input to subtask $\tau^{(i+1)}$. Continue until $i=N$.

\BlankLine
\textbf{Step 4: Goal verification} \\
If $s^{(N)} \models G$, the multi-agent VLA workflow succeeds.
\end{algorithm}

\paragraph{High-Level Planner.}
LLMs and VLMs excel at high-level planning due to their large-scale training, which equips them with extensive world knowledge and commonsense reasoning skills. One may leverage this capability to generate abstract plans; as $\pi_{0.5}$~\cite{pi02024} used a VLM to encode text and image and guide the low-level action generation expert. In multi-agent scenarios, a high-level planner may also serve as an orchestrator, assigning tasks to the low-level action experts.

\paragraph{Low-Level Action Expert.}
A low-level action expert generates a sequence of actions to achieve each subgoal, that may have been generated by the high-level planner. It can be implemented either as a diffusion-based or an autoregressive model trained on real robot action datasets, such as, Open-X-Embodiment~\cite{open_x_embodiment_rt_x_2023} or DROID~\cite{khazatsky2024droid}. The low-level action expert may also take input from multiple modalities---e.g., vision, text, and touch among others.
% ---as modality-specific tokens from corresponding encoders.

\paragraph{Reasoning before Actions.}  
Rather than directly sampling low-level actions, one can introduce an intermediate reasoning layer that grounds each subtask $\tau^{(i)}$ in structured, interpretable guidance. This paradigm---termed \emph{reasoning before actions}---encourages the policy to articulate how a subgoal should be achieved prior to generating the corresponding motor commands~\cite{sun2024emmaxembodiedmultimodalaction,zawalski2024robotic}. 

Consider the robot as a \textit{delivery driver} tasked with delivering a package to a destination (the goal). The driver first decomposes the journey into subgoals: ``Reach Main Street'', ``Cross the river bridge'', ``Arrive at the central library''. For each subgoal, the driver reasons about the appropriate strategy: to reach the Main Street, ``follow Elm Road until the first intersection''; to cross the river bridge, ``merge onto the highway ramp''. These reasoning steps ground the low-level driving actions---such as turning the wheel left, pressing the accelerator, or braking---which are executed in sequence to realize the subgoal.  

Formally, a subtask $\tau^{(i)}$ is realized through a sequence of low-level actions
\[
A^{(i)} = \bigl[a^{(i)}_{1},\,a^{(i)}_{2},\dots,a^{(i)}_{T_i}\bigr], 
\quad 
a^{(i)}_{j} \sim \pi_{k_i}\!\left(s_{t_j}^{(i-1)},\,\tau^{(i)}\right),
\]
with state transitions
\[
s_{t_{j}+1}^{(i-1)} \;=\; \mathcal W\!\left(s_{t_j}^{(i-1)},\,a^{(i)}_{j}\right).
\]

Before producing $A^{(i)}$, the system generates a reasoning trace
\[
r^{(i)} = f_{\text{reason}}\!\left(\tau^{(i)},\,s_{t_0}^{(i-1)}\right),
\]
which provides both linguistic guidance (e.g., ``move up by 22 steps, then left by 23 steps'') and semantic rationale (e.g., ``close the gripper around the pot-lid’s handle to secure the grasp''). This reasoning trace constrains and grounds the subsequent action generation.  

\noindent\textbf{Example.} Consider a high-level plan for cooking:  
\begin{itemize}[itemsep=0pt, leftmargin=*, wide, labelwidth=0pt, labelindent=0pt, parsep=0pt, topsep=0pt]
    \item \textit{Plan}: Poise above pot-lid $\rightarrow$ Grasp pot-lid $\rightarrow$ Lift lid.  
    \item \textit{Subtask} $\tau^{(i)}$: Grasp pot-lid.
    \item \textit{Reasoning} $r^{(i)}$: ``Align the gripper with the pot-lid’s handle and close it firmly to ensure a stable grasp.''  
    \item \textit{Low-level actions} $A^{(i)}$: move up $22$ steps, move left $23$ steps, move down $23$ steps, close gripper.  
\end{itemize}

By explicitly generating $r^{(i)}$ before $A^{(i)}$, the policy links linguistic intent with motor execution, yielding more robust and interpretable action generation.

\subsection{Improving Perception by Spatial Understanding}

As discussed in \cref{sec:chal-perc}, most VLA models are limited in their spatial understanding due to limited depth perception, being reliant on two-dimensional RGB-only images as input. To remedy this, the VLAs could be trained to understand and reason with the precise depth information obtained from depth-gauging cameras, LIDAR sensors, or multi-camera setup, should they be available. Approaches using depth estimation or depth-aware perception tokens~\cite{lee2025molmoactactionreasoningmodels,spatialvla2025}, could only be used as a fallback option. 

Unfortunately, most VLM backbones of VLAs lack inherent depth-aware spatial understanding. Thus, VLMs could be fine-tuned with both real and synthesized RGB-D frame data to inculcate depth-aware spatial understanding. Locate 3D~\cite{arnaud2025locate3drealworldobject} framework could be used to synthesize such RGB-D frames from RGB frames. Such data could be sourced from many Visual QA datasets, but the constituent queries may not necessarily require depth understanding. Thus, one may leverage Locate 3D to insert new objects from a predefined set into the frames and construct new QA pairs requiring depth awareness. Alternatively, powerful video generation platforms and world models like Veo3 could be leveraged to construct diverse scenes and form QA pairs from them. The RGB-D frames could be constructed from the moving camera positions---farther objects move slower than the closer objects in the frame.

Following a pre-training regimen on synthetic depth-containing visual data, the depth-aware VLM could be trained on synthetically constructed RGB-D frames from the Open-X-Embodiment~\cite{open_x_embodiment_rt_x_2023} dataset. For standard depth-devoid RGB frames, one can keep a separate expert that generates estimated-depth-aware encodings of the frames like MolmoAct and SpatialVLA.

%\paragraph{Proprioception.}

\subsection{Universal Action Representation}
As highlighted in \Cref{sec:chal-gen}, cross-robot action generalization remains a central challenge in VLA research. One promising direction is to learn unified atomic representations of actions. For instance, \citet{zheng2025universal} proposed learning universal atomic actions via a codebook and then decoding them into robot-specific actions through a decoder. Their results suggest that this approach significantly reduces the effort and data required to adapt to new robots. However, it still falls short of achieving true zero-shot generalization. We envision that this gap could be addressed by teaching VLA models about a new robot’s action space through prompting---analogous to few-shot prompting that teaches LLMs new tasks, generally referred to as in-context learning. Realizing this capability, however, would require a fundamental shift in the pre-training paradigm of VLA models.

\subsection{Modeling World Dynamics}
 %\jw{"World Modeling" could be too broad --- people may argue that "Spatial Understanding" can also be part of world modeling. Maybe something like "Modeling World Dynamics" or "Predictive/Dynamics Modeling"?}
World models are integral to robotics frameworks, as they model the cause-and-effect relationships essential for predicting the outcomes of actions. Recent studies have highlighted that current visual-language models (VLMs) often face challenges in executing plans. These difficulties may primarily arise from interpreting the current state, anticipating desired outcomes, and grounding both in the physical world to determine the required actions. These challenges underscore the need for explicit world models.

The two major approaches to world models are
\begin{enumerate}[itemsep=0pt, leftmargin=*, wide, labelwidth=0pt, labelindent=0pt, parsep=0pt, topsep=0pt]
    \item \textbf{Generative modeling:} In this approach, a world model $\mathcal{W}$ takes a current state $s$ and an action $a$, and predicts the resulting next state: $s' \;\leftarrow\; \mathcal{W}_\theta(s, a)$.
    Such a model can be initialized from a pretrained VLM or LLM. To enforce self-consistency, additional objectives—such as predicting the action that causally connects two states—can be included. These objectives are often formulated contrastively, promoting correct state predictions while demoting incorrect ones, and can be implemented via specialized prediction heads to avoid interfering with next-state prediction. Flow-based architectures are one possible instantiation of this paradigm.
    
    \item \textbf{Embedding prediction:} This approach, proposed by Yann LeCun and realized in models such as JEPA~\cite{dawid2023introductionlatentvariableenergybased} and V-JEPA~\cite{bardes2024revisitingfeaturepredictionlearning}, predicts the masked patches of visual frames as pre-training objective to learn the embeddings. V-JEPA-2~\cite{assran2025vjepa2selfsupervisedvideo} extends this idea by post-training a hierarchical predictor: given the current observation/visual context and an action $a_t$, predict the latent embedding of the future frame. 
    Such approach circumvents explicit pixel-level reconstruction by abstracting away irrelevant visual details, allowing efficient long-horizon prediction. This could be particularly suitable for robotics where reasoning about high-level state changes is more important than exact pixel reproduction.
\end{enumerate}

\subsection{Data Synthesis with Visual Generative Models}
Large Language Models are trained on trillions of tokens that are readily available from internet-scale corpora. In contrast, collecting data at scale for Embodied AI models, including Vision-Language-Action (VLA) systems, is far more challenging, as it typically requires costly robot teleoperation. A common workaround is to use simulators and then mix simulated trajectories with a limited amount of real robot data for sim-to-real transfer. However, achieving robust sim-to-real generalization remains a long-standing challenge in robotics.

Recent advances in video generative models provide a promising alternative. Such models can synthesize videos of robots accomplishing tasks in diverse environments, where the environments themselves can be generated from textual prompts at scale. This opens new opportunities for VLA research, but introduces a key limitation: generated videos do not contain explicit robot action sequences. To address this, one can extract \emph{latent actions}~\cite{yang2025comolearningcontinuouslatent} from video via world modeling, as detailed below.

\paragraph{Joint Learning from Video and Robot Data.}
We introduce a latent action variable \(z\) for world model \(\mathcal{W}\). Concretely, we train an encoder \(E\) to infer $z_t = E(s_t, s_{t+1})$
from consecutive states, and optimize the world model to predict dynamics $s_{t+1} =\mathcal{W}_\theta(s_t,\,z_t)$.
Since ground-truth future states are available in both simulated and video data, we can compute a reconstruction loss and backpropagate to jointly train \(E\) and \(\mathcal{W}\). This formulation sidesteps the need for manual action annotations by grounding policy learning in a self-supervised latent space.

However, the latent representation \(z\) does not directly correspond to the real robot action space. To bridge this gap, we leverage datasets with ground-truth robot actions $a_t$ that can be used to optimize \(E\) and $\mathcal W$, while jointly learning $z_t$. This joint learning can then be realized through
\begin{enumerate}[itemsep=0pt, leftmargin=*, wide, labelwidth=0pt, labelindent=0pt, parsep=0pt, topsep=0pt]
    \item Minimizing the distance between the latent representation \(z_t\) and the true action $a_t$.
    \item Encoding $a_t$ into the latent space and ensuring $\mathcal{W}$ predicts $s_{t+1}$ to either $z_t$ or $a_t$ as action input.
\end{enumerate}
This hybrid training scheme allows video-derived latent actions to be aligned with real robot control signals, enabling scalable pretraining on synthetic video while remaining grounded in executable action spaces.

\paragraph{Video Generative Models and World Models as Simulators.}  
Video generative models and world models such as V-JEPA-2 offer a powerful means of producing simulated data for training embodied agents. These models can be fine-tuned to predict future states conditioned on the current state and a sequence of actions. To generate large-scale simulated datasets, one can initialize from random frames and sample actions from a predefined distribution to produce corresponding future states. An evaluator can then filter the generated states to retain only those that satisfy specific subgoals, ensuring data quality. The resulting cleaned dataset can subsequently be used to train low-level action experts.

\subsection{Post-Training}
In LLM research, significant gains in System-II level intelligence and task performance have been achieved through extensive post-training using reinforcement learning. The key idea is that the model explore the solution space by generating multiple rollouts. A reward model then evaluates these rollouts based on criteria such as task completion, efficiency, and optimality. The reward signals are used to update the policy, enabling the model to gradually improve its action selection and favor strategies that achieve higher rewards. This process effectively combines exploration of possible actions with guided learning from feedback, allowing the model to discover increasingly effective behaviors. Extending this paradigm to Visual-Language-Action (VLA) models faces a fundamental challenge: how can we define and provide reward signals for these models? 

A straightforward approach might involve running the VLA model $N$ times to generate action sequences $A^{(i)}$, executing them in a simulator, and then using evaluators to assess the outcomes and assign rewards. However, this approach depends on the availability of highly accurate and efficient robot-specific simulators—a requirement that is often difficult and costly to meet.

Recent advances in world models and video generative models offer a promising alternative. These models can serve as implicit reward estimators by predicting the consequences of actions and evaluating whether desired subgoals are achieved. Leveraging such learned models as reward functions could enable scalable post-training of VLA policies without the need for fully engineered simulators, providing a practical path forward for reinforcement learning in embodied settings.

\paragraph{Improving the Action Expert through Rewards.}  
Given $N$ rollouts from the action expert, action-conditioned world dynamics models can be used to predict the resulting future states. These states can then be compared to the ground-truth subgoal states using a suitable metric to obtain reward signals for the action expert. Alternatively, LLMs can assess if the predicted outcomes satisfy the intended subgoals, providing feedback to guide policy improvement. Note that one can also use different other rewards, both verifiable or unverifiable. Preference optimization techniques, such as, Direct Preference Optimization (DPO)~\cite{rafailov2023direct}, RL, and GRPO (Group Reward Preference Optimization)~\cite{shao2024deepseekmathpushinglimitsmathematical}, can be adopted for improving the action expert.

\paragraph{Safety Check through Evaluators.}
Ensuring safety and task compliance is critical to deploying VLA models. One approach is to leverage action-conditioned world models or specialized evaluators---or reward models for post-training---as virtual guardrails to simulate the proposed actions and detect potential safety violations or deviations from task objectives or trigger human interventions. If a safety or compliance issue is identified, the VLA model can replan to reach the goal, while avoiding unsafe behaviors.
Such evaluators are especially important in multi-agent settings, where interactions between autonomous VLA agents can introduce emergent risks. By incorporating these safety checks, the system can maintain reliable and controlled behavior even in complex, collaborative environments.

\section{Conclusion}
Vision-Language-Action (VLA) models are central to the development of Embodied AI. In this work, we revisit several key challenges in this area—such as multimodal sensing and perception, and cross-robot action generalization—that we believe will shape future research directions. In addition, we propose an exploratory framework aimed at addressing these challenges.

\section*{Acknowledgement}
This work was supported in part by A*STAR SERC CRF funding to C.T.  This work is also supported
by the National Research Foundation, Singapore, under its National Large Language
Models Funding Initiative (AISG Award No:
AISG-NMLP-2024-005), NTU SUG project
\#025628-00001:Post-training to Improve Embodied AI Agents.
\bibliography{aaai2026,custom}

\end{document}